\documentclass[letterpaper, 10 pt, conference]{template/ieeeconf}
\IEEEoverridecommandlockouts                              

\usepackage{graphicx}
\usepackage{caption}
\usepackage{amsmath,amssymb,enumerate}

\usepackage{amsthm}
\usepackage{mathtools}
\usepackage{breqn}
\usepackage{algorithm, algorithmicx, algpseudocode}

\usepackage{blindtext}
\usepackage{gensymb}
\usepackage{xparse}
\usepackage{lipsum}
\usepackage{mathrsfs}
\usepackage[mathscr]{euscript}
\usepackage{times}
\usepackage[noadjust]{cite} 
\usepackage{multicol}
\usepackage[caption=false,font=footnotesize]{subfig}
\usepackage{amsfonts}
\usepackage[T1]{fontenc}
\usepackage{textcomp}
\usepackage{balance}
\usepackage{soul}
\usepackage{multirow}
\usepackage{booktabs}
\usepackage[usenames,dvipsnames,svgnames,table, xcdraw]{xcolor}

\usepackage[colorlinks=true,pdfpagemode=UseNone,citecolor=black,linkcolor=black,urlcolor=BrickRed]{hyperref}
\usepackage{comment}

\newtheorem{theorem}{Theorem}

\newtheorem{remark}{Remark}

\newtheorem{definition}{Definition}

\renewcommand{\thefigure}{\arabic{figure}}

\newcommand{\transpose}{\mathsf{T}}

\newcommand{\m}{\mathop{\mathrm{m}}}
\newcommand{\rad}{\mathop{\mathrm{rad}}}

\newcommand{\Hz}{\mathop{\mathrm{Hz}}}

\title{\LARGE \bf Fully Proprioceptive Slip-Velocity-Aware State Estimation for Mobile Robots via Invariant Kalman Filtering and Disturbance Observer}

\author{Xihang Yu, Sangli Teng, Theodor Chakhachiro, Wenzhe Tong, Tingjun Li, Tzu-Yuan Lin\\
        Sarah Koehler, Manuel Ahumada, Jeffrey M. Walls, and Maani Ghaffari
\thanks{Toyota Research Institute provided funds to support this work. Funding for M. Ghaffari was in part provided by NSF Award No. 2118818.}%
\thanks{X. Yu, S. Teng, T. Chakhachiro, W. Tong, T. Li, T.Y. Lin and M. Ghaffari are with the University of Michigan, Ann Arbor, MI 48109, USA. {\tt\small\{xihangyu,sanglit,teochiro,wenzhet\}@umich.edu},
{\tt\small\{tingjunl,tzuyuan,maanigj\}@umich.edu}}%
\thanks{ S. Koehler, M. Ahumada, and J. Walls are with Woven Planet. \texttt{\{sarah.koehler,manuel.ahumada\}@woven-planet.global}, \texttt{jeff.walls@woven-planet.global}}
}

\begin{document}

\maketitle
\thispagestyle{empty}
\pagestyle{empty}

\begin{abstract}
This paper develops a novel slip estimator using the invariant observer design theory and Disturbance Observer (DOB). The proposed state estimator for mobile robots is fully proprioceptive and combines data from an inertial measurement unit and body velocity within a Right Invariant Extended Kalman Filter (RI-EKF). By embedding the slip velocity into $\mathrm{SE}_3(3)$ matrix Lie group, the developed DOB-based RI-EKF provides real-time velocity and slip velocity estimates on different terrains. Experimental results using a Husky wheeled robot confirm the mathematical derivations and effectiveness of the proposed method in estimating the observable state variables. Open-source software is available for download and reproducing the presented results.
\end{abstract}

\section{Introduction}
\label{sec:introduction}

For mobile robots, state estimation is the problem of determining a robot's position, orientation, and velocity for a fixed world or body frame~\cite{barfoot2017state}. This problem is vital for many robotics tasks such as control~\cite{gong2021one, agrawal2022vision, teng2022error} or trajectory planning~\cite{gan2022energy, teng2022lie, Teng-RSS-23}. 
The filtering approach takes low computational resources and often operates at a high frequency~\cite{hartley2020contact,lin2021legged, gao2022invariant,van2022eqvio}. 
One interesting class of estimators is the Invariant Extended Kalman Filter (InEKF). Matrix Lie groups provide natural (exponential) coordinates that exploit symmetries of the space~\cite{chirikjian2011stochastic,long2013banana,barfoot2014associating,hall2015lie,barfoot2017state}. The theory of invariant observer design is based on the estimation error being invariant under the action of a matrix Lie group, leading to constant linearizations, linear observability analysis, and convergence property for the deterministic group-affine systems~\cite{barrau2017invariant,barrau2018invariant,hartley2020contact}. Hence, we adopt this framework here.

One challenge for robot state estimation is the slip incurred by uneven, deformable, or low-friction terrains. Once a slip happens, it violates the fixed contact point assumption, thus resulting in drifts. Another interesting aspect of detecting slip is stability control. If one could detect slip accurately, the controller can use this information to maintain robot stability~\cite{wallace2019receding,bledt2018contact}. 

In this work, we develop a novel slip estimator using a Right Invariant Extended Kalman Filter (RI-EKF). The proposed model estimates the body slip velocity in the world frame by treating it as a bias in the velocity measurements. Given the encoder-inertial data on wheeled platforms, the slip velocity can be estimated in a fully proprioceptive manner. In particular, this work has the following contributions.
\begin{enumerate}
    \item A real-time RI-EKF state and slip estimator for wheeled mobile robots using only onboard encoder-inertial measurements.
    \item A slip event observer as a binary classifier using Chi-square statistical test. 
    \item Experimental results using a Husky robot on multiple terrains to verify the mathematical derivations and validate the effectiveness of the proposed state estimator.
    \item Open source software is available for download at \href{https://github.com/UMich-CURLY/slip\_detection\_DOB}{\small https://github.com/UMich-CURLY/slip\_detection\_DOB}
\end{enumerate}
\begin{figure}[t]
    \centering
    \includegraphics[width= .99\columnwidth]{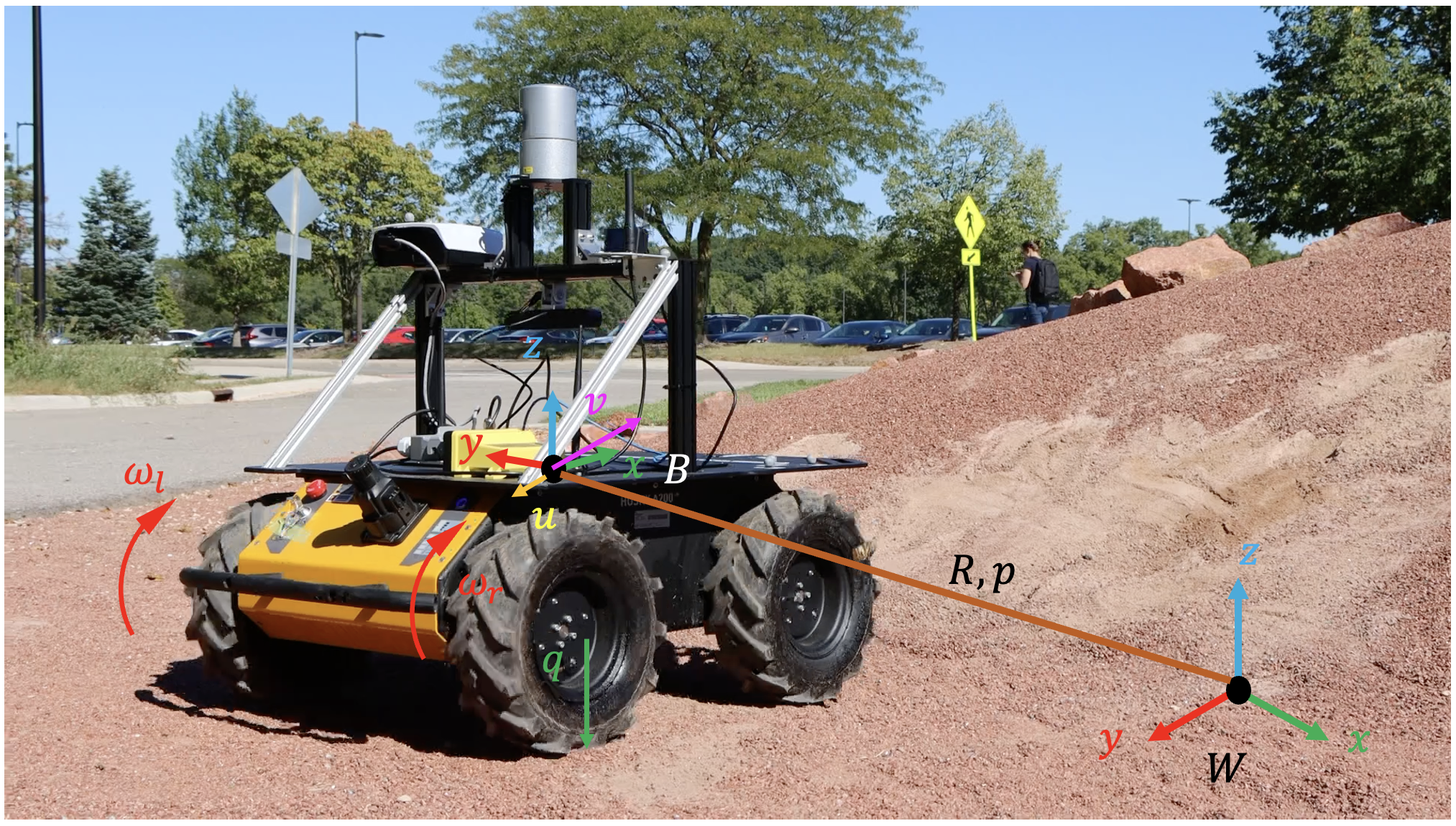}
    \caption{Experimental wheeled robot platform Husky. The world frame $W$ is a fixed frame. We define the body frame $B$ as $x$ pointing forward, $y$ pointing to the left of the vehicle, and $z$ derived from the right-hand rule. We obtain angular velocity measurements $\omega_l$ and $\omega_r$ from the wheel encoders. Using an inertial measurement unit and wheel encoder measurements, the proposed method estimates the body velocity, slip velocity, robot orientation, and position, visualized as $v$, $u$, $R$, and $p$, respectively. The observability analysis shows that, given the employed sensor configuration, the absolute position and yaw angle (robot heading) are unobservable while the remaining state variables are observable.}
    \label{husky}
\end{figure}


\section{Related Work}
\label{sec:literature}

Direct measurements, e.g., using biomimetic tactile sensors \cite{james2018slip,li2018slip}, can provide slip detection; however, specialized hardware is required, and such sensors can be too fragile for mobile robots. Therefore, algorithmic methods based on limited sensing data are attractive solutions for slip detection.

\subsection{Model-based methods}
The work of~\cite{Pico2022} designs a slip controller on a six-wheel mobile robot to perform obstacle climbing maneuvers. Slip is detected by comparing the wheel velocity from the forward kinematics with the body velocity estimated by the visual-inertial system. A slip event is classified as true if the normalized difference is larger than a fixed threshold. The work of~\cite{Malinowski2022} fuses multiple models for wheel slip prediction in a nonlinear observer. For instantaneous slip detection, this work requires the visual measurement or the wheel motor current. Then a more complicated slip detection model is incorporated to predict the slip events. 
These methods rely on accurate body velocity estimation from a camera or privileged information of the friction profile, whereas the proposed work here is fully proprioceptive.

Other than fusing kinematics or visual data, more work used the dynamics model of the vehicles for state estimation. \cite{piyabongkarn2008development} fuses the simplified vehicle dynamics, kinematics models, and tire models to estimate the \textit{slip angle}, i.e., the angle between the vehicle heading direction and actual velocity. The slip angle is crucial for friction coefficient and slip estimation. The work of~\cite{imsland2006vehicle} uses the full dynamics of the vehicle and tire models to estimate the longitudinal and lateral speed of the vehicles using a nonlinear observer. Based on~\cite{imsland2006vehicle}, the work of \cite{grip2008nonlinear} applied an update law to the friction parameters so that the nonlinear observer could adapt to varying friction conditions. However, the works of \cite{piyabongkarn2008development, imsland2006vehicle, grip2008nonlinear} rely on the vehicle dynamics model and assumes known terrain inclinations or certain friction parameters. In our method, we only assume the kinematics model and do not have any assumption on the terrain inclination or friction. 

\subsection{Learning-based methods}
\textcolor{black}{We have seen increasing interest and great improvements in using learning-based methods for slip detection/estimation tasks as shown in survey \cite{lopez2021machine,gonzalez2019characterization}.} The work of~\cite{omura2017wheel} uses the Support Vector Machine (SVM) for slip detection. An in-wheeled sensor is used to measure the normal force and contact angle. \cite{gonzalez2018deepterramechanics, gonzalez2018slippage} explored various learning algorithms to estimate slip for wheeled robots. These methods model the slip estimation task as a classification problem. The main drawback of the supervised learning-based method is that ground truth is hard to obtain. In~\cite{omura2017wheel}, the ground truth is hand labeled, which is laborious and prone to error. Although \cite{gonzalez2018deepterramechanics} does not rely on hand labeling data, the experiment is conducted in a laboratory environment using a single wheel. The ground truth is obtained by measuring the wheel's angular velocity and the carriage pulley's angular velocity. The work of~\cite{gonzalez2018slippage} uses other sensors, such as GPS for ground truthing, which suffers from outages (i.e., indoors or shaded areas). The works of~\cite{bouguelia2017unsupervised,kruger2019estimating} detect slip using unsupervised learning. However, the slip detection is modeled as classification, leading to inaccessible slip velocities.

The reinforcement learning-based system identification has also drawn attention recently. \cite{yu2017preparing} trained a network in the simulator to estimate the unknown parameters of mechanical models for control. \cite{kumar2021rma, kumar2022adapting} encode the environmental factors to latent spaces, enabling online system identification for varying environments. However, the learned high-dimensional latent space lacks interpretability. In this work, we develop a low-dimensional slip model that can be estimated recursively without the need for offline training.

\subsection{Filtering-based methods}

The contact-aided InEKF \cite{hartley2020contact} fuses the foot contact state, forward kinematics, and Inertial Measurement Unit (IMU) measurements, assuming the contact point is fixed in the world frame. For legged robot state estimation on slippery terrain, \cite{bloesch2013state} applied the Unscented Kalman Filter (UKF) to fuse the end-effector velocity and a Chi-square detection to discard unreliable measurements when the foot is not static. The discarding method enables more robust performance but also reduces the number of measurements. The work of~\cite{teng2021legged} fuses visual-based velocity measurement with leg kinematics and inertial measurements on slippery terrain. Incorporating visual information can compensate for unreliable measurements from leg kinematics. The above methods either assume static contact points or treat slip events as outliers. \textcolor{black}{The works of \cite{yang2022wheel,kilic2022proprioceptive,rogers2012aiding} use the Extended Kalman Filter (EKF) in slip estimation tasks. \cite{yang2022wheel} estimate slip ratio and sinkage on deformable terrain. \cite{kilic2022proprioceptive} can only detect slip events but leave slip velocity estimation inaccessible. Similarly to our work, \cite{rogers2012aiding} models wheel slip as compensation of velocity, but the slip velocity is still not fully observable and needs GPS to calibrate the velocity compensation.}


The Disturbance Observer (DOB) estimates the disturbance by explicitly representing it as a new state variable~\cite{chen2015disturbance, han2009pid}. The DOB does not assume accurate dynamics of the disturbances but uses simple dynamics with higher generalizability. In this work, we model the slip velocity as an exponentially decaying variable, i.e., an autoregressive process, which is simple yet can capture the core dynamics of slippage. Compared to previous works, we address the problem of slip velocity estimation using a lightweight DOB that is fully proprioceptive and computationally attractive.

\section{Mathematical Preliminary and Notation}
\label{sec:prelimina}

 We assume a matrix Lie group denoted $\mathcal{G}$ and its associated Lie Algebra denoted $\mathfrak{g}$. 
 Let $(\cdot)^\wedge:\mathbb{R}^{\mathrm{dim} \mathfrak{g}} \to \mathfrak{g}$ be the isomorphism that takes elements of the tangent space of $\mathcal{G}$ at the identity to the corresponding matrix representation so that 
 $\exp:\mathbb{R}^{\mathrm{dim} \mathfrak{g}} \to \mathcal{G}$, is computed by $\exp(\xi) = \exp_m(\xi^\wedge)$, where $\exp_m(\cdot)$ is the usual exponential of $n \times n$ matrices. A process dynamics evolving on the Lie group with the state at time $t$, $X_t \in \mathcal{G}$, is denoted by
$\frac{d}{dt} X_t = f_{u_t}(X_t), $
and $\hat{X}_t$ is used to denote an estimate of the state. 
\begin{definition}[Right-Invariant Error] 
The right-invariant error between two trajectories $X_t$ and $\hat{X}_t$ is:
\begin{equation} \label{eq:invariant_error}
\begin{split}
\eta_t^r &= \hat{X}_t X_t^{-1} = (\hat{X}_t L) (X_t L)^{-1} ,
\end{split}
\end{equation}
where $L$ is an arbitrary element of the group.
\end{definition}
The following two theorems are the fundamental results for deriving an InEKF.
\begin{theorem}[Autonomous Error Dynamics \cite{barrau2017invariant}] \label{theorem:autonomous_error_dynamics}
A system is group affine if the dynamics, $f_{u_t}(\cdot)$, satisfies:
\begin{equation} \label{eq:group_affine}
f_{u_t}(X_1 X_2) = f_{u_t}(X_1) X_2 + X_1 f_{u_t}(X_2) - X_1 f_{u_t}(I_d) X_2,
\end{equation}
for all $t>0$ and $X_1, X_2 \in \mathcal{G}$. Furthermore, if this condition is satisfied, the right-invariant error dynamics are trajectory independent and satisfy:
\begin{alignat*}{2}
\frac{d}{dt} \eta_t^r &= g_{u_t}(\eta_t^r) \quad \text{where} \quad
g_{u_t}(\eta^r) &&= f_{u_t}(\eta^r) - \eta^r f_{u_t}(I_d) .
\end{alignat*}
\end{theorem}
The group identity element is denoted $I_d \in \mathcal{G}$; we use $I$ for a $3 \times 3$ identity matrix and $I_n$ for the $n \times n$ case. 

Define $A_t$ to be a $\mathrm{dim} \mathfrak{g} \times \mathrm{dim} \mathfrak{g}$ matrix satisfying $g_{u_t}(\exp(\xi)) := (A_t \xi)^\wedge + \mathcal{O}(\lVert \xi \rVert^2).$
For all $t \ge 0$, let $\xi_t$ be the solution of  the linear differential equation
\begin{equation}
\label{eq:LTVODE}
\frac{d}{dt} \xi_t = A_t \xi_t.
\end{equation}

\begin{theorem}[Log-Linear Property of the Error \cite{barrau2017invariant}] \label{theorem:log_linear_error}
Consider the right-invariant error, $\eta_t$, between two trajectories (possibly far apart). For arbitrary initial error $\xi_0 \in \mathbb{R}^{\mathrm{dim} \mathfrak{g}}$, if
\mbox{$\eta_0 =\exp(\xi_0)$}, then for all $t\ge 0$, 
$\eta_t = \exp(\xi_t)$;
that is, the nonlinear estimation error $\eta_t$ can be exactly recovered from the time-varying linear differential equation \eqref{eq:LTVODE}.
\end{theorem}

Utilizing a right-invariant observation model \cite{barrau2017invariant}, we get an autonomous linearized observation model and innovation.
\begin{equation} \label{eq:invariant_observations}
\begin{alignedat}{2} 
Y_t &= X_t^{-1} b + n_t \quad &&\text{(Right-Invariant Observation)},
\end{alignedat}
\end{equation}
where $b$ is a constant vector, and $n_t$ is a vector of Gaussian noise.
Finally, for any $X \in \mathcal{G}$ the adjoint map, \mbox{$\mathrm{Ad}_X:\mathfrak{g} \to \mathfrak{g}$}, is a linear map defined as \mbox{$(\mathrm{Ad}_X \xi)^\wedge = X \xi^\wedge X^{-1}$}. The similarity transformation via the adjoint map allows a change of frame in the Lie algebra.

\section{System models}
\label{sec:method}
In this section, we introduce the kinematics model of the differential wheeled robot and derive the dynamics of wheel slip velocity. For a quantity $\zeta$, we use $\tilde{\zeta}$ to denote its measurement and $\hat{\zeta}$ to indicate its estimation. 
\subsection{IMU model}
The state of the IMU can be represented by its orientation $R \in \mathrm{SO}(3)$, velocity $v \in \mathbb{R}^3$, and position $p \in \mathbb{R}^3$ in the world frame~\cite{hartley2020contact}. 
The IMU measures the acceleration ${{a}}\in \mathbb{R}^3$ and angular velocity ${{\omega}}\in \mathbb{R}^3$ in the IMU frame. Consider the white Gaussian noise in the gyroscope ${w}_{\omega} \sim \mathcal{N}({0_3}, {\Sigma^{\omega}})$ and accelerometer ${w}_{a} \sim \mathcal{N}({0_3}, {\Sigma^{a}})$, the IMU measurements can be represented by $\tilde{{\omega}}={\omega}+{w}_{\omega}$, $\tilde{{a}}={a}+{w}_{a}$. Given the IMU measurements, its unbiased dynamics can be represented as
\begin{equation}
     \dot{{R}} = {R}\left(\tilde{{\omega}}-{w}_{\omega}\right)^\wedge, \quad \dot{{v}} = {R}\left(\tilde{{a}}-{w}_{a}\right)+{g},	\quad \dot{{p}} = {v},
\end{equation}
where ${g}$ is the gravity vector. For bias augmentation and discretization, we follow the same approach as that of~\cite{hartley2020contact}.

\subsection{Differential wheel robot model}
\label{subsec:diff_wheel}
The angular velocities of the left and right wheels around their axles, i.e., ${{\omega}}_l,{{\omega}}_r \in \mathbb{R}$, can be measured by the encoders. Let $q \in \mathbb{R}$ be the known wheel radius. Assuming zero velocity at the contact points, the forward velocities of the wheel radius centers are $\tilde{v}_{xl} = {{\omega}}_l q$ and $\tilde{v}_{xr} = {{\omega}}_r q$.

Due to the nonholonomic constraints of the wheels, we assume that the robot has zero lateral and vertical velocity. This pseudo-measurement model has been used previously in \cite{dissanayake2001aiding,brossard2020ai}. For the differential wheel robot, there is one additional constraint in forward velocity; the forward velocity is the average of velocities for left and right wheel radius centers. Combining all measurements, we form the observation model without slip estimation as 
\begin{equation}
    {y}_{ENC} := \operatorname{vec}\left( \frac{1}{2}\left( \tilde{v}_{xl} + \tilde{v}_{xr}\right), 0, 0 \right) = R^{\top}v .
	\label{eq_mea_pseudo}
\end{equation}

\subsection{Wheel slip model}
The differential wheel kinematics in \ref{subsec:diff_wheel} assumes the non-slip condition, which may not be valid when a slip event happens. Inspired by the DOB, we explicitly model the slip velocity as a new state variable. Let $u \in \mathbb{R}^3$ be the slip velocity (disturbance) in the world frame. We append $u$ to \eqref{eq_mea_pseudo} and we have the new observation model:
\begin{equation}
	{y}_{ENC} = R^{\top} v + R^{\top} u + w_y ,
	\label{eq_mea_pseudo_dob}
\end{equation}
where $\mbox{$w_y \sim \mathcal{N}(0_{3}, W_y)$}$ is a lumped zero-mean white Gaussian noise term that takes into account the uncertainty in kinematics model, slip events, and encoders. 

The philosophy of DOB is to use a simple model to represent the dynamics of disturbance. One widely adopted method is to model the disturbance as a constant \cite{chen2015disturbance, han2009pid}. However, this model does not match reality as the slip event dissipates energy. Instead, we use an autoregressive process to model the slippage. Thus, the dynamics of slip compensation in the world frame is modeled as follows.
\begin{equation}
\label{eq:dynamics_slip}
    \dot{u} = - \alpha u + R {w}_u,  \quad {w}_{u}  \sim \mathcal{N}({0_3}, {\Sigma^{u}}) ,
\end{equation}
\textcolor{black}{where the slip velocity's decay rate is denoted by real positive value $\alpha > 0$. We treat $w_u$ as a single noise term that reflects the imprecision of the slip model and other stochastic effects. This model is consistent with natural observations, as discussed in Remark~\ref{slip_dynamic}. The slip velocity is represented by an exponential function, as depicted in \eqref{eq:dynamics_slip}. A similar result has been derived in \cite{vanderkop2022novel} to represent friction.} 

\begin{remark}
\label{slip_dynamic}
 Suppose the robot is slipping on icy ground. After gaining an initial momentum and without any external forces other than friction, the robot will slip forward with a decreasing velocity in the world frame and eventually stop. This observation inspires the above autoregressive process.
\end{remark}

\section{The Invariant Extended Kalman Filtering}
\label{sec:inekf}

In the following, we use the extended pose matrix Lie group $\mathrm{SE}_3(3)$. For more detail about this group and its structure, we refer to~\cite{hartley2020contact}.
\subsection{State representation on Lie group}

Define the state matrix on the Lie group $\mathrm{SE}_3(3)$ as:
\begin{equation}\label{eq:state}
{X}=\left[\begin{array}{cccc}
{R} & {v} & {p} & {u}\\
{0}_{3,3} &  & I & 
\end{array}\right] \in \mathrm{SE}_3(3).
\end{equation}

\textcolor{black}{The continuous-time noisy process dynamics of the augmented state $X \in \mathrm{SE}_3(3)$ is given by}
\begin{align}
\nonumber \frac{\mathrm{d}}{\mathrm{dt}} {X} & = f_{u_{t}}({X}) - {X} w^\wedge \\
&:= \left[\begin{array}{cccc} 
{R}\tilde{{\omega}}^\wedge & {R}\tilde{{a}}+{g} & {v} & -\alpha u\\
{0}_{3,3} & 0_{3,1} & 0_{3,1} & 0_{3,1} \\
\end{array}\right] - {X} w^\wedge ,
\label{eq:continuous_dynamic}
\end{align}
\textcolor{black}{with} $w := \operatorname{vec}(w_w, w_a,0_{3,1},w_u)$ is the zero-mean white Gaussian noise process evolving in the Lie algebra.
Then we have the right-invariant error $\eta^r \in \mathrm{SE}_3(3)$ and its first-order approximation as:
\begin{align}
\nonumber {\eta^r} & =\hat{{X}} {X}^{-1} \\
\nonumber & =\left[\begin{array}{lccc}
\hat{{R}} {R}^{\transpose} & \hat{{v}}-\hat{{R}} {R}^{\transpose} {v} & \hat{{p}}-\hat{{R}} {R}^{\transpose} {p} & \hat{{u}}-\hat{{R}} {R}^{\transpose} {u} \\
{0}_{1 , 3} & 1 & 0 & 0 \\
{0}_{1 , 3} & 0 & 1 & 0 \\
{0}_{1 , 3} & 0 & 0 & 1 \\
\end{array}\right] \\
& = \exp(\xi) \approx\left[\begin{array}{cccc}
{I}+{\xi}_{R}^{\wedge} & {\xi}_{v} & {\xi}_{p} & {\xi}_{u}\\
{0}_{1 , 3} & 1 & 0 & 0 \\
{0}_{1 , 3} & 0 & 1 & 0 \\
{0}_{1 , 3} & 0 & 0 & 1
\end{array}\right].
\end{align}
\textcolor{black}{with} \mbox{${\xi}:=\operatorname{vec}\left({\xi}_R,{\xi}_v, {\xi}_p,{\xi}_u \right) \in \mathbb{R}^{\mathrm{12}}$} and $\xi^{\wedge} \in \mathfrak{se}_3(3)$ is the equivalent noise in the Lie algebra. 



\subsection{InEKF Propagation Step}

\textcolor{black}{The deterministic system dynamics $f_{u_t}(\cdot)$ can be shown to satisfy the group affine property (\ref{eq:group_affine}). Therefore, according to Theorem~\ref{theorem:autonomous_error_dynamics}, the right-invariant error dynamics will be autonomous and evolve independently of the system’s state:}
\begin{align}
\nonumber \frac{d}{dt} \eta_t^r & = f_{u_t}(\eta_t^r) - \eta_t^r f_{u_t}(I_d) + \left( X_t w_t^\wedge X_t^{-1} \right)  {\eta_t^r} \\
& := g_{u_t}(\eta_t^r) + \left( X_t w_t^\wedge X_t^{-1} \right) {\eta_t^r},
\end{align}
\textcolor{black}{where the second term is from the additive noise. The derivation follows the results of Barrau and Bonnabel~\cite{barrau2017invariant}.}

\textcolor{black}{Theorem \ref{theorem:log_linear_error} states that the dynamics of the $g_{u_t}$ can be fully captured by its linearization in the Lie algebra~\footnote{We note that the exact result only holds for the deterministic system. Nevertheless, the noisy system performs well in practice as shown in previous works~\cite{barrau2017invariant,barrau2018invariant,hartley2020contact,lin2021legged}}; thus, by deriving $A_t$ from $g_{u_t}(\exp(\xi_t)) := (A_t \xi_t)^\wedge + \mathcal{O}(\lVert \xi_t \rVert^2)$, we have the linear error dynamic system }
\begin{equation}
\frac{\mathrm{d}}{\mathrm{d} t}
\xi_{t}={A}\xi_{t}
+
\operatorname{Ad}_{\hat{{X}}_{t}}{w}_{t}
\end{equation}
where
\begin{equation}
	{A}=\left[\begin{array}{ccccc}{0}_{3} & {0}_{3}& {0}_{3}& {0}_{3}\\
	{g}^\wedge& {0}_{3}& {0}_{3}& {0}_{3}\\ 
	{0}_{3} & {I}_{3}& {0}_{3}& {0}_{3}\\
	{0}_{3}& {0}_{3}& {0}_{3}& -\alpha I_3\\ 
    \end{array}\right]
	\label{eq_A},
\end{equation}
and $\operatorname{Ad}_{\hat{{X}}_{t}}$ is the group adjoint map~\cite{hartley2020contact}. The covariance of the augmented right invariant error dynamics is computed by solving the Riccati equation
$
\frac{\mathrm{d}}{\mathrm{d} t} {P}_{t}={A} {P}_{t}+{P}_{t} {A}^\transpose+\hat{{Q}}_{t},
$
where $\hat{{Q}}_{t}=\operatorname{Ad}_{\hat{{X}}_{t}} \operatorname{Cov}\left({w}_{t}\right) \operatorname{Ad}_{\hat{{X}}_{t}}^\transpose$.


\subsection{Right-Invariant Correction Step}

We can re-write~\eqref{eq_mea_pseudo_dob} as a right-invariant observation model,
	${y}=\hat{X}^{-1} b+n$, where $y := \operatorname{vec}\left(\frac{1}{2} (\tilde{v}_l + \tilde{v}_r),-1,0,-1\right)$, $b := \operatorname{vec}\left( {0}_{1 \times 3},-1,0,-1 \right)$, and $n := \operatorname{vec}\left( w_y,0,0,0 \right)$.
Then we compute the right-invariant measurement Jacobian, $H$, using ${H}{\xi}^r = -({\xi}^r)^\wedge {b} $ as
${H}=\left[{0}_{3},I,{0}_{3},I\right].$

Following RI-EKF methodology~\cite{barrau2017invariant,hartley2020contact}, we can compute the innovation covariance $S=H P H^{\transpose}+W_y$ and the filter gain $K=P H^{\transpose} S^{-1}$ .
Let $\Pi := \left[I,0_3\right]$ be a column selection matrix. We then compute an updated state and covariance, using \mbox{$\hat{X}^+ = \exp\left( K \Pi \left( \hat{X}y - b \right) \right)\hat{X}$} and 
\mbox{${P}^{+} = ({I}_{12}-{KH})P$}, respectively.

\subsection{Observability Analysis}

Because the error dynamics are log-linear (c.f., Theorem \ref{theorem:log_linear_error}), we can determine the unobservable states of the filter without nonlinear observability analysis \cite{barrau2015non}. Noting that the linear error dynamics matrix in our case is time-invariant and nilpotent (with a degree of three), the discrete-time state transition matrix, ${\Phi}=\exp _{m}\left({A} \Delta t\right)$, is 
$$
{\Phi} = \left[\begin{array}{cccc}
{I} & {0} & {0} & {0}\\
({g})^{\wedge} \Delta t & {I} & {0}& {0}\\
\frac{1}{2}({g})^{\wedge} \Delta t^{2} & {I} \Delta t & {I} & {0}\\
{0} & {0} & {0} & \exp(-\alpha I \Delta t)\\
\end{array}\right].
$$
It follows that the discrete-time observability matrix is 
$$
\mathcal{O}=\left[\begin{array}{c}
{H} \\
{H} {\Phi} \\
{H} {\Phi}^{2} \\
\vdots
\end{array}\right]=\left[\begin{array}{cccc}
{0} & {I} & {0}& {I} \\
(g)^{\wedge} \Delta t & {I} & 0 & \exp(-\alpha I \Delta t) \\
2(g)^{\wedge} \Delta t & {I}& 0 & \exp(-\alpha I \Delta t)^2 \\
\vdots & \vdots & \vdots & \vdots
\end{array}\right].
$$
The second last column (i.e., one matrix column) of the observability matrix are zeros, indicating the absolute position of the robot, which is unobservable. Because the gravity vector only has a $z$ component, a rotation about the gravity vector (yaw) is also unobservable. Moreover, we note that slip disturbances are fully observable except for $\alpha = 0$. This case would not occur since we have assumed $\alpha > 0$ in (\ref{eq:dynamics_slip}).

\subsection{Chi-square hypothesis testing for slip detection}
To classify slip events, we use the Chi-square hypothesis testing. The application of the Chi-square test resembles its use in \cite{bloesch2013state}. However, unlike \cite{bloesch2013state}, where outlier observations are filtered out, we classify slip events. Note that we have modeled the dynamics of slip velocity $u$ as an autoregressive process in~\eqref{eq:dynamics_slip}. Therefore, $u$ is exponentially stable, and its covariance has a bounded steady state value for a bounded noise input. From the property of exponential stability, the mean value of $u$ converges to a steady state $\mathbb{E}[u]=:\bar{u} = 0$, and the covariance converges to a steady state value $\operatorname{Cov}\left(\bar{u}\right) =: \Sigma_{\bar{u}}$. The steady state covariance of $u$ is unknown and does not necessarily equal the time-varying covariance estimated by the filter. As such, we test if $u$ matches an empirical steady-state distribution $\mathcal{N}(0,\Sigma_{\bar{u}})$ using a Chi-square test, where we manually tune $\Sigma_{\bar{u}}$ in this work to be a constant covariance matrix. We compute the Chi-square statistics with $n=3$ degrees of freedom as:
\begin{equation}
    r := (u- \bar{u})^\transpose \Sigma_{\bar{u}}^{-1}(u- \bar{u}) 
      = u^\transpose \Sigma_{\bar{u}}^{-1}u \sim {\chi}^{2}_{n=3}.
\end{equation}
\textcolor{black}{If $r$ is larger than a threshold given a confidence level, we consider the robot state as slipping; otherwise, non-slipping. Note that the conventional slip ratio approach to representing slippage may not be entirely applicable in our scenario as we do not presume that slipping only takes place in the longitudinal axis. Our proposed Chi-square statistical method allows for the possibility of using data-driven techniques to learn the covariance $\Sigma_{\bar{u}}$ in the future. }

\textcolor{black}{In Section \ref{sec:results}, we used a grid search to determine the steady state covariance. We tested $I$, $0.1\cdot I$, $0.01\cdot I$, $0.001\cdot I$, $0.0001\cdot I$ and $0.00001\cdot I$ and $\Sigma_{\bar{u}}=0.001\cdot I$ turned out to have the best performance. For a probability value of $0.80$ with $3$ degrees of freedom, a threshold value of $4.642$ is computed from the Chi-square distribution table. Statistically, it suggests that we have 80\% confidence that any slip velocity samples that fall outside of this threshold are outliers, indicating a slip event has occurred. On the other hand, if the observed value is smaller than this threshold, it is considered an inlier and classified as a non-slip event. Developing a data-driven classifier for improved performance is an attractive future work; however, the proposed statistical test is computationally efficient and easy to implement.} 


\section{Experimental Results}
\label{sec:results}

We conduct two experiments to validate the proposed methods on a Husky robot, as shown in Fig.~\ref{husky}. In the first experiment, the robot moves over a highly slippery soap area back and forth. In the second experiment, the robot climbs onto a sandy hill twice. With these two test sequences, we focus on evaluating the performance of the proposed method on velocity and orientation estimation, as well as slip detection. For the tracking results of the InEKF on the Husky platform, see~\cite{ghaffariprogress2022}. 

As we test our method in the outdoor area without a motion capture system, we use a state-of-the-art visual SLAM system, ORB-SLAM3~\cite{campos2021orb}, to serve as a proxy for the ground truth velocity. The images used in ORB-SLAM3 are streamed from a ZED camera mounted on the robot. The camera is used for ground truth generation only. The proposed method is entirely proprioceptive, relying solely on the data from an IMU and wheel encoders as the sensors. In our experimental setup, the IMU and wheel encoder measurements are received at a frequency of $200 \Hz$ and $13.5 \Hz$, respectively. \textcolor{black}{ To prevent the varying sampling rates of the IMU and encoder from impacting the estimation results, we utilize a First-In-First-Out data queue. When new data is added to the queue, the system checks if it's IMU or encoder data. If it's IMU data, the propagation step is applied. Conversely, if it's wheel encoder data, the correction step is applied.} In addition to the proposed method, we run the InEKF without the DOB as a baseline algorithm. The measurement noise statistics and initial covariance estimates are provided in Table \ref{parameters1}. We fixed hyperparameters in all experiments (including both baseline and proposed filters) for better comparison. 

\begin{table}
\centering
\caption{Noise Parameters and initial Covariances. All values show the standard deviations, and covariances are spherical.}
\resizebox{\columnwidth}{!}{
\begin{tabular}{l|lll|l}
\cline{1-2}\cline{4-5}
Measurement Type    & Noise std. dev.    &  & State Variable     & Initial Covariance  \\
\cline{1-2}\cline{4-5}
Gyroscope    & 0.1 rad/s       &  & Robot Orientation  & 0.03 rad            \\
Accelerometer & 0.1 $\mathrm{m/s^{2}}$ &  & Robot Velocity     & 0.01 m/s              \\
Gyroscope Bias      & 0.001 rad/s       &  & Robot Position     & 0.01 m             \\
Accelerometer Bias  & 0.001 $\mathrm{m/s^{2}}$   &  & Robot Slip Velocity     & 0.01 $\mathrm{m/s}$         \\
Wheel Encoder Vel  & 0.1 m/s            &  & Gyroscope Bias     & 0.0001 rad/s        \\
Disturbance Process  & 5 m/s           &  & Accelerometer Bias & 0.0025 $\mathrm{m/s^2}$             \\
\cline{1-2}\cline{4-5}
\end{tabular}}
\label{parameters1}
\end{table}

\subsection{DOB as State Estimator}
\label{DOB_state_estimator_result}

We use the observation model described in \eqref{eq_mea_pseudo_dob} to correct the predicted states. However, instead of setting a constant measurement covariance, $W_y$, we make $W_y$ adaptive based on the estimated slip velocity, i.e., 
\begin{equation}
W_y = W_{0} \exp(\lVert u \rVert) .
\label{cov_adap}
\end{equation}
Here, $W_{0}$ is the initial covariance that corresponds to the covariance when no slip is present. When the robot is suffering from severe slip, the measurement covariance increases exponentially with respect to the estimated slip velocity.

\begin{table}[t]
\caption{RMSE comparison between the baseline (InEKF) and the proposed filter (InEKF w/ DOB) in the experiment of Sec. \ref{DOB_state_estimator_result}. We compute the RMSE error during $44\sec-57\sec$ with respect to each rotation (in $\rad$) and velocity (in $\m/\sec$) axis.}
\centering
\label{table:rmse1}
\begin{tabular}{lcccccc}
\hline
State & yaw & pitch & roll & $v_x$ & $v_y$ & $v_z$  \\
\hline
InEKF & 0.098 & 0.018 & 0.027 & 0.155 & \textbf{0.052} & 0.039 \\
InEKF w/ DOB & 0.098 & \textbf{0.008} & \textbf{0.015} & \textbf{0.100} & 0.055 & \textbf{0.036} \\
\hline
\end{tabular}
\end{table}

\begin{figure}[t]
    \centering
    \vspace{3mm}
    \subfloat{\includegraphics[width=      0.8\columnwidth]{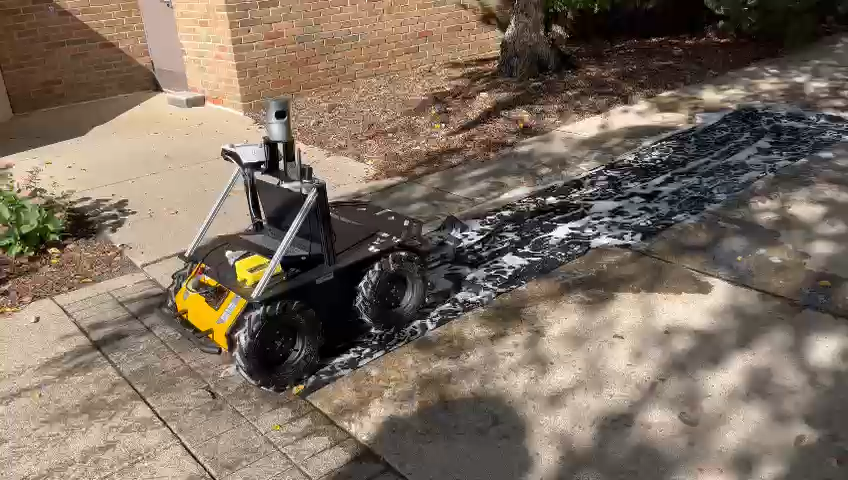}}\\
    \caption{The Husky robot is navigating on a highly slippery soap-covered tape ground (black areas).}
    \label{soap}
\end{figure}

\begin{figure}[t]
    \centering
    \includegraphics[width=      1\columnwidth]{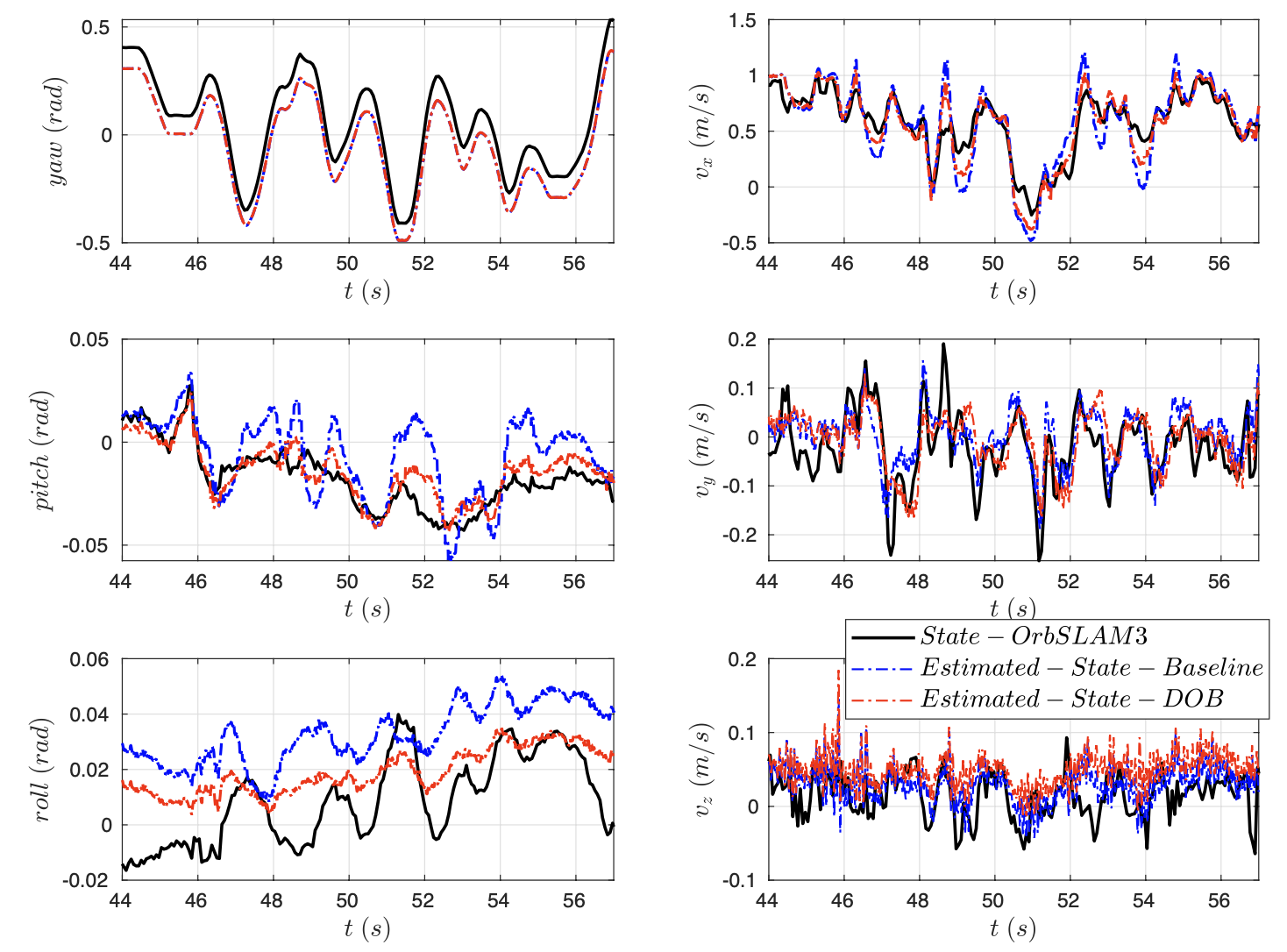}
    \caption{Yaw, pitch, and roll angles of the main body are depicted in the left three figures. Body Velocities are depicted in the right three figures. Black lines: ground truth from ORB-SLAM3 system. Blue dashed: plain InEKF without DOB. Red dashed: InEKF with DOB. Note that the ORB-SLAM3 algorithm only provides pose estimation. The velocity is the result of differentiating the estimated pose.}
    \label{rot_bodyvel}
    \vspace{-3.5mm}
\end{figure}

In the first experiment, Husky moves on a soapy surface back and forth, as demonstrated in Fig.~\ref{soap}. Fig.~\ref{rot_bodyvel} plots the estimated orientation and velocity in this sequence. Note that $v_y$ and $v_z$ are not zeros due to the actual slippage. We focus on the period from $44 \sec$ to $57 \sec$, in which the robot moves through the slippery area.\footnote{The rosbag of this dataset can be found on our GitHub page.} For attitude estimation, the yaw angle is unobservable, as pointed out in observability analysis.

Root Mean Square Error (RMSE) comparison between the baseline and the proposed filter is presented in Table~\ref{table:rmse1}. The proposed algorithm performs better than the baseline pitch and roll estimation, which are essential for vehicle stability, especially in off-road scenarios. The disturbance observer provides accurate attitude estimation in the slippery area. 
For body velocity estimation, the proposed algorithm outperforms the baseline algorithm~\cite{ghaffariprogress2022} in the $x$-axis (dominant dimension of the signal here), and there is a slight improvement in the $z$ direction. However, the baseline performs slightly better in $y$-axis velocity estimation. We conjecture this is due to inappropriate covariance adaptation by \eqref{cov_adap}. 

\begin{figure}[t]
    \centering
    \vspace{3mm}
    \includegraphics[width=      0.8\columnwidth]{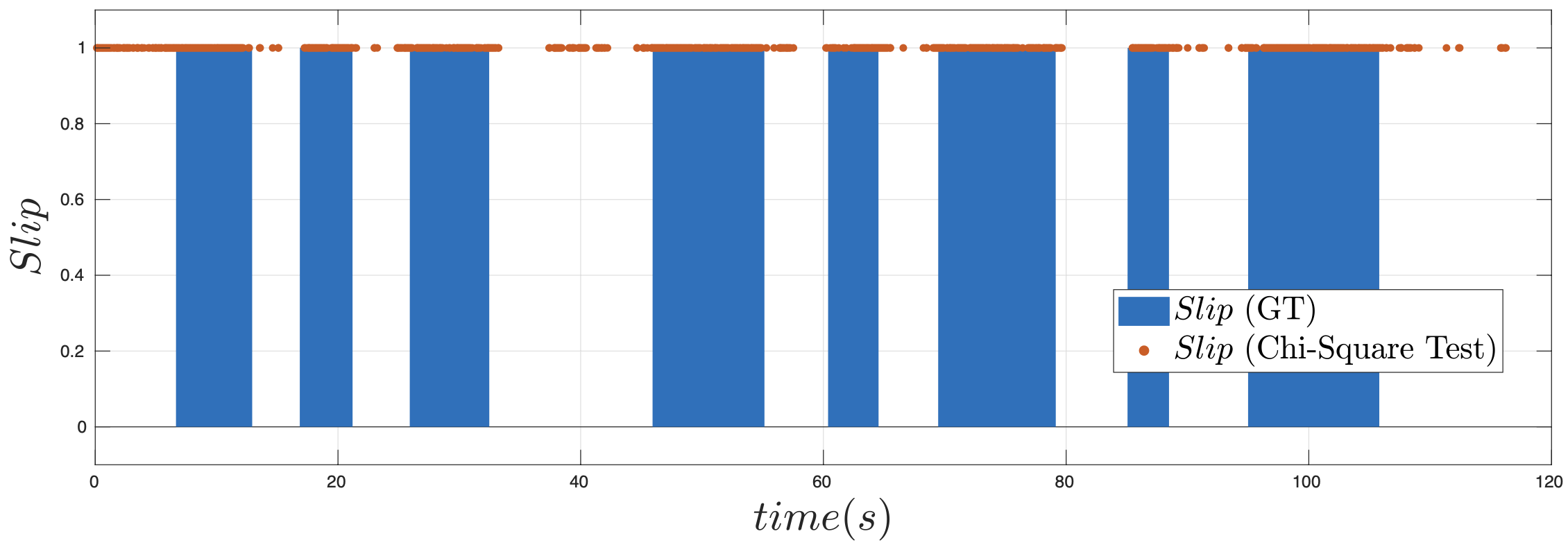}
    \caption{Slip state estimation from sequence over slippery soap terrain. We use 1 for slipping and 0 for non-slipping. Red dots: Detected slip event during which the robot is slipping. Blue bars: Ground truth during which the robot is slipping.}
    \label{slip_State}
\end{figure}

\begin{figure*}[t]
    \centering
    \subfloat{\includegraphics[width=      2\columnwidth]{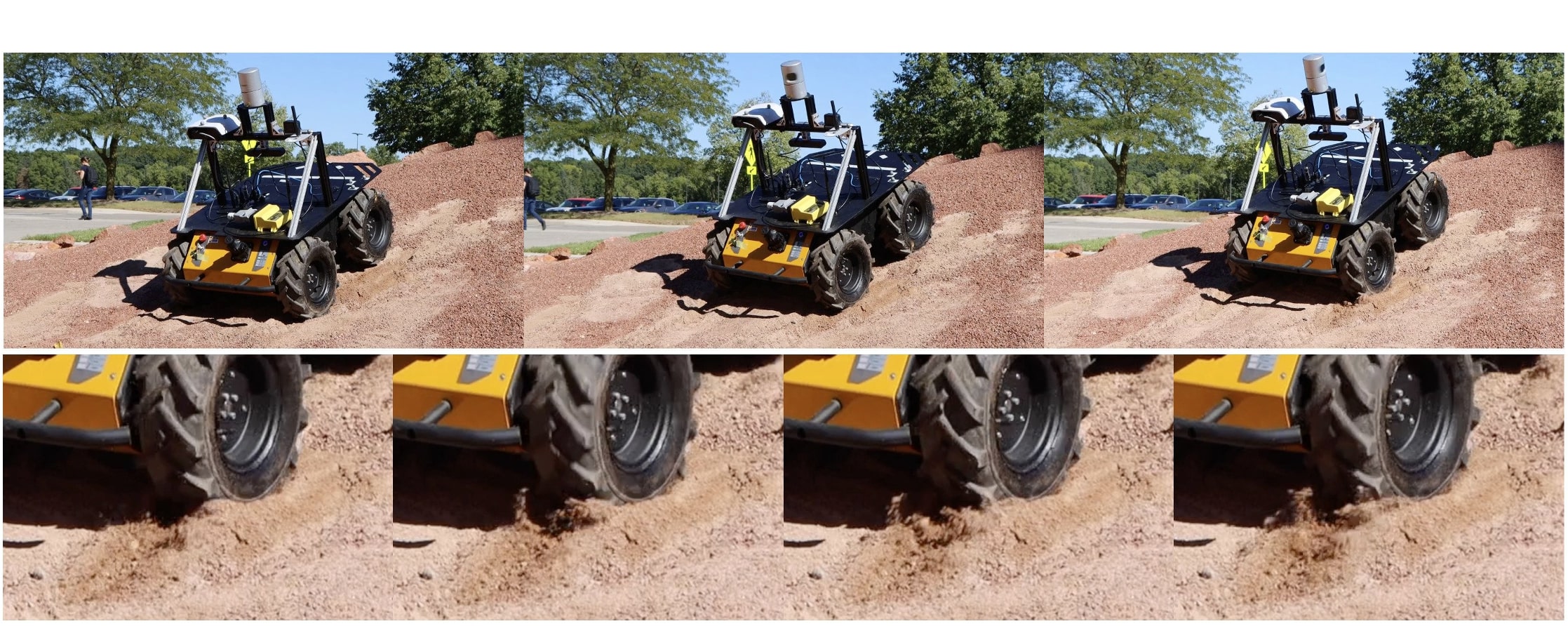}}\\
    \caption{A sequence of snapshots illustrating the substantial slip that is occurring during the experiment. If looking at the contact points of the right rear wheel with the ground, one can observe that sands are rolling behind when the wheel is rotating.} 
    \label{husky_mars_hill_image}
    \vspace{-4.75mm}
\end{figure*}
 
\subsection{DOB as Slip Event Observer}
\label{DOB_slip_observer_result}
Slip detection is crucial for the safety and efficiency of autonomous systems \cite{kilic2022proprioceptive}. Disturbance observer also serves as a slip-aware filter. We use a Chi-square test to classify slip states as binary events. \textcolor{black}{We manually label ground truth data by observing the slip event in the recorded video.} When the robot turns around or moves through the soap area, we consider it a slip event, while on normal ground, we consider it non-slip. We show the result of the Chi-square test in the 120-second sequence in Fig.~\ref{slip_State}. We set $\chi^2$ confidence level at 80\% and the corresponding threshold as $4.642$.

\begin{table}[t]
\caption{The average performance of the proposed filter (InEKF w/ DOB) in slip estimation.}
\centering
\label{table:slip_est}
\setlength{\tabcolsep}{3.5mm}{
\begin{tabular}{lccc}
\hline
Experiment &  FPR &  FNR &  Accuracy  \\
\hline
Artificial slippery terrain & $20.33\%$ & $25.60\%$ & $77.32\%$ \\
Deformable sand terrain  & $59.14\%$ & $7.78\%$ & $65.76\%$ \\
\hline
\end{tabular}}
\end{table}

\textcolor{black}{We compute the False Positive Rate (FPR), False Negative Rate (FNR), and accuracy as follows. 
$$FPR = \frac{FP}{FP+TN} \quad \text{and} \quad FNR = \frac{FN}{FN+TP},$$
$$Accuracy = \frac{TP+TN}{TP+TN+FP+TN}.$$
We achieve about 77.32\% accuracy (Fig.~\ref{soap}) as shown in Table~\ref{table:slip_est}. Furthermore, there is 20.33\% FPR, i.e., detected slip but in reality not a slip event, and 25.60\% FNR, i.e., slip events that are not detected.}



\begin{table}[t]
\caption{RMSE comparison between the baseline (InEKF) and the proposed filter (InEKF w/ DOB) in the experiment of Sec. \ref{mars_hill_result}. }
\centering
\label{table: rmse2}
\footnotesize
\begin{tabular}{lcccccc}
\hline
State & yaw & pitch & roll & $v_x$ & $v_y$ & $v_z$  \\
\hline
InEKF & 0.046 & \textbf{0.056} & 0.035 & 0.305 & 0.092 & 0.140 \\
InEKF w/ DOB & 0.046 & 0.057 & 0.035 & \textbf{0.277} & \textbf{0.087} & \textbf{0.077} \\
\hline
\end{tabular}
\end{table}

\subsection{Experiment on Highly Deformable Terrain}
\label{mars_hill_result}
\begin{figure}[t]
    \centering
    \includegraphics[width=      1\columnwidth]{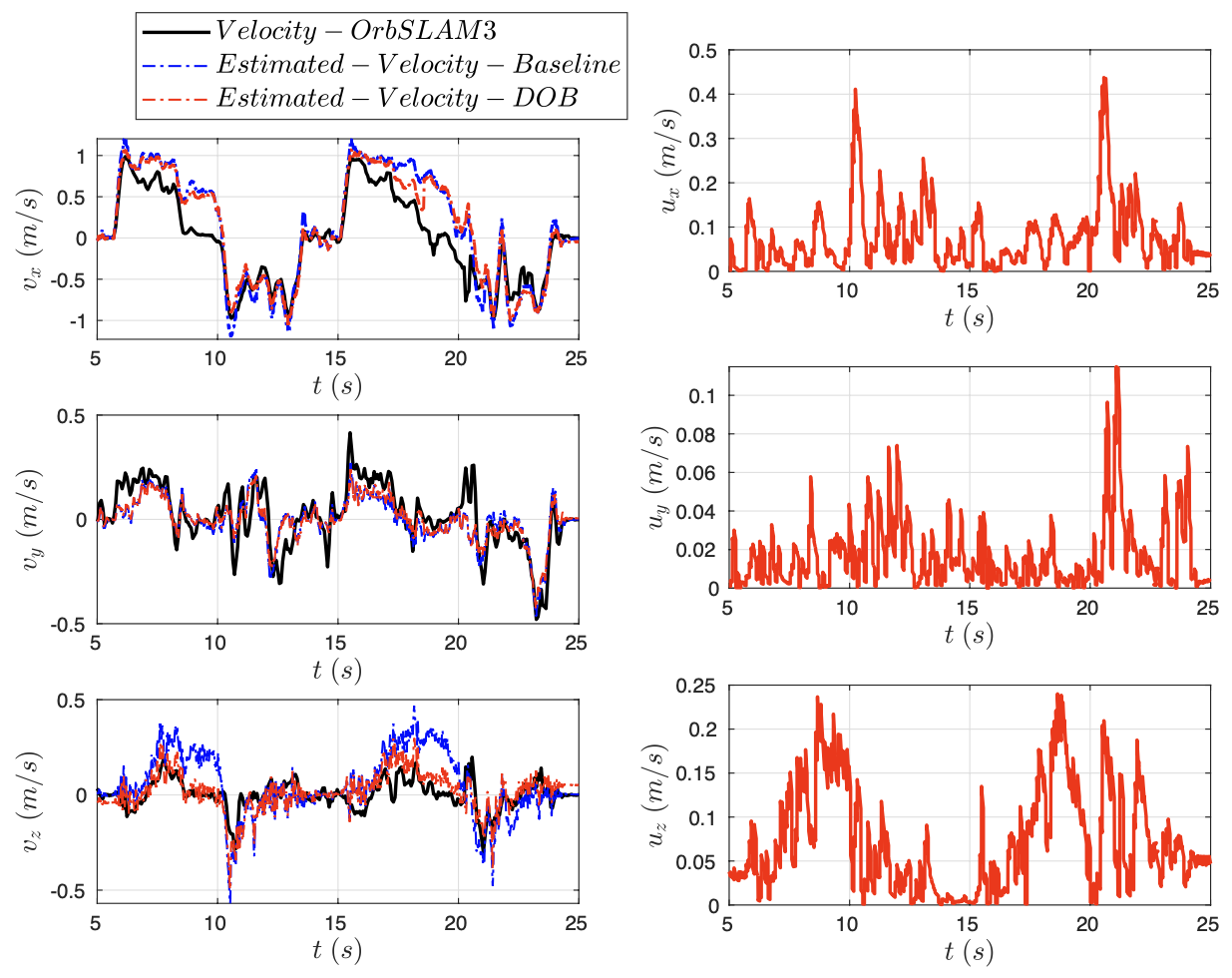}
    \caption{Estimated velocity and disturbance for the hill sequence. Note that the estimated velocity and disturbance are represented in the world frame. The left three figures plot the estimated velocities from the proposed method (Red), the baseline (Blue), and the ground truth velocity from ORB-SLAM3 (Black). The right three figures present the estimated slip velocities from the proposed method. The two peaks in $x$ and $z$ directions represent severe slipping when the robot is on the slope of the hill. 
    }
    \label{vel_up_to_hill3}
\end{figure}

The second experiment is conducted outdoors on a deformable terrain. As illustrated in Fig.~\ref{husky_mars_hill_image}, the Husky was driven twice straight up to a hill covered with sand. Fig.~\ref{vel_up_to_hill3} shows this estimated velocity and disturbance of this sequence in the world frame. \textcolor{black}{ In Fig.(\ref{husky}) The $x$-axis is pointing forward, $y$-axis is pointing left and $z$-axis is pointing upward.} Compared to the baseline method, the proposed algorithm provides more accurate velocity estimations when Husky is experiencing severe slipping (seconds 8-12 and seconds 18-22). The quantitative results of this sequence are shown in Table~\ref{table: rmse2}. The proposed filter outperforms standard InEKF in all three velocity directions while keeping similar results in orientation. Slip velocity estimation is shown in the right three subplots in Fig.~\ref{vel_up_to_hill3}. The two peaks in the $x$ and $z$ directions represent severe slipping when the robot is on the slope of the hill. We observe that the peaks of the $z$-axis appear at second 8 and second 18 because at those moments robot is at the highest slope on the hill. Peaks of the $x$-axis appear later at second 11 and second 21 when the robot returns to the low-slope valley. To validate the proposed method as a slip detector, we run the Chi-square test in Sec.~\ref{DOB_slip_observer_result}. Quantitative results are shown in Table~\ref{table:slip_est}. We achieve an accuracy of 65.76\% for sand terrain.

\textcolor{black}{In Table \ref{table: rmse2}, there is no clear improvement observed in the attitudes prediction of InEKF w/DOB when comparing sand terrain to an artificial soap terrain. For the sand terrain experiment, the FPR is found to be much higher than the FNR. This indicates that the estimator used for predicting slip events is over-confident. This discrepancy in the FPR and FNR and the degradation of the state estimator can be attributed to the use of the same set of hyperparameters, such as the decaying rate $\alpha$, for slip classification in both the soap and sand terrains. In reality, these parameters should be adjusted for different terrains. A valuable extension of this work would be to adapt the hyperparameters in real time using data-driven techniques.}

\textcolor{black}{The ground truth for slip is determined through manual labeling of tire tracks. When the tire track is apparent, it is classified as a non-slip condition, whereas when the tire track is indistinct, it is identified as a slip condition. This labeling method can cause noise and a motion capture system can be a better way to label ground truth events.} 

\section{Discussion and Limitations}
\label{sec:limitations}
In this work, we model the slip velocity of the whole body frame within a RI-EKF framework. Currently, the decaying rate $\alpha$ should be tuned by hand. However, the value of the hyperparameter should reflect the ground-tire friction coefficient (i.e., for more slippery terrain, the alpha should be smaller). It is known that the knowledge of the friction dynamics can improve the vehicle's model fidelity~\cite{park2017discrete}. Our proposed filter might be used in friction coefficient identification. The covariance matrix of encoder measurements is set heuristically in the InEKF and updated adaptively by a manually designed function. This can cause inaccuracy sometimes, as in the $v_y$ estimation in Fig.~\ref{rot_bodyvel}. In the future, we wish to use a data-driven method to estimate the covariance of the current measurements~\cite{brossard2020ai}. This approach will allow the state estimator to fully exploit the information in disturbance estimation. Moreover, the steady-state distribution for the disturbance is also set heuristically in the Chi-Square test. This value could potentially also be learned by a neural network. Finally, the developed slip disturbance observer could be integrated into other platforms, such as legged robots, where the body velocity measurements are used~\cite{wisth2020preintegrated}.

\section{Conclusion}
\label{sec:conclusion}
We developed a lightweight filtering-based slip velocity estimation method that only uses encoder and IMU data and works in real-time on distinct terrains. In addition, a Chi-square hypothesis testing approach is proposed for detecting slip events. We tested the proposed algorithm on a Husky wheeled robot and demonstrated better performance than a standard InEKF. The discussed limitations motivate future studies in this area of robot state estimation. The proposed slip velocity estimator is a low-cost addition to the existing invariant EKF and has the potential to be widely adopted on various robotic platforms. 

{\small
\balance
\bibliographystyle{bib/IEEEtran}
\bibliography{bib/strings-abrv,bib/ieee-abrv,bib/refs}
}

\end{document}